\documentclass[letterpaper]{article} 
\usepackage{aaai25}  
\usepackage{times}  
\usepackage{helvet}  
\usepackage{courier}  
\usepackage[hyphens]{url}  
\usepackage{graphicx} 
\usepackage{natbib}  
\usepackage{caption} 
\usepackage{algorithm}
\usepackage{algorithmic}

\usepackage{newfloat}
\usepackage{listings}
\DeclareCaptionStyle{ruled}{labelfont=normalfont,labelsep=colon,strut=off} 
\floatstyle{ruled}
\newfloat{listing}{tb}{lst}{}
\floatname{listing}{Listing}
\usepackage{pifont}
\usepackage{color, colortbl}

\title{\makebox[28pt][l]{\hspace{0.7em}\raisebox{-0.25\height}{\includegraphics[height=1.8em]{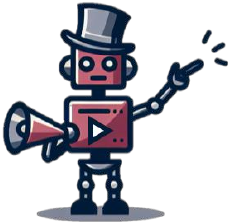}}} AutoDirector: Online Auto-scheduling Agents for Multi-sensory Composition}
\author {
    Minheng Ni\textsuperscript{\rm 1}\equalcontrib,
    Chenfei Wu\textsuperscript{\rm 1}\equalcontrib,
    Huaying Yuan\textsuperscript{\rm 1},
    Zhengyuan Yang\textsuperscript{\rm 2},\\
    Ming Gong\textsuperscript{\rm 3},
    Lijuan Wang\textsuperscript{\rm 2},
    Zicheng Liu\textsuperscript{\rm 2},
    Wangmeng Zuo\textsuperscript{\rm 4},
    Nan Duan\textsuperscript{\rm 1},
}
\affiliations {
    \textsuperscript{\rm 1}Microsoft Research Asia \quad \textsuperscript{\rm 2}Microsoft Azure AI\\
    \textsuperscript{\rm 3}Microsoft Search \& Distribution\quad 
    \textsuperscript{\rm 4}Harbin Institute of Technology\\
    \texttt{\{t-mni, chewu, v-huayuan, zhengyang, migon, lijuanw, zliu, nanduan\}@microsoft.com}
}

\usepackage{bibentry}

\usepackage{multirow}
\usepackage{makecell}
\usepackage{hhline}
\usepackage{multicol}
\usepackage{booktabs}
\usepackage{amssymb}

\begin{document}

\twocolumn[{
\renewcommand\twocolumn[1][]{#1}
\maketitle
\centering

\captionsetup{type=figure}
\includegraphics[width=17cm]{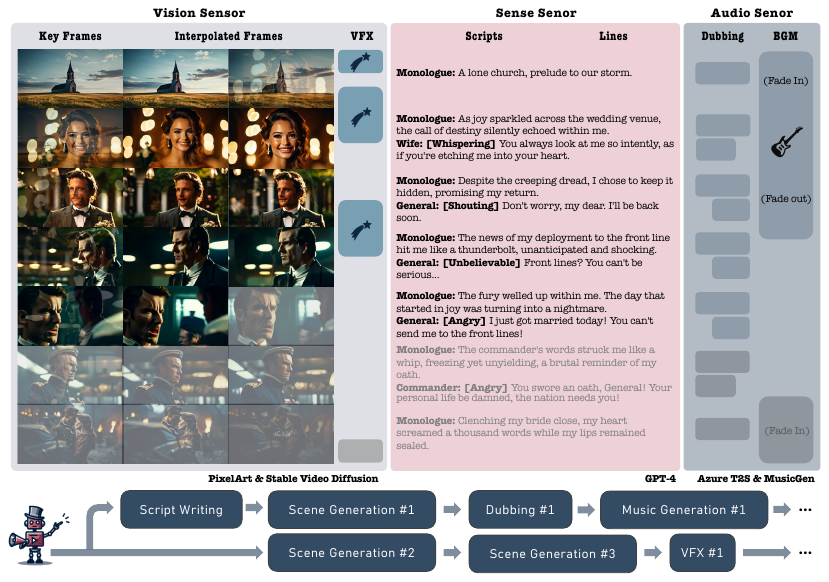}
\captionof{figure}{\textbf{Multi-sensory movie generation results of AutoDirector.} The multi-sensory composition process includes scriptwriting, shooting, scoring, dubbing, and special effects. AutoDirector effectively integrates these elements to produce high-quality movies.}
\label{fig:case}
\vspace{1em}
}
]

\begin{abstract}
With the advancement of generative models, the synthesis of different sensory elements such as music, visuals, and speech has achieved significant realism. However, the approach to generate multi-sensory outputs has not been fully explored, limiting the application on high-value scenarios such as of directing a film. 
Developing a movie director agent faces two major challenges: 
(1) Lack of parallelism and online scheduling with production steps: In the production of multi-sensory films, there are complex dependencies between different sensory elements, and the production time for each element varies.
(2) Diverse needs and clear communication demands with users: Users often cannot clearly express their needs until they see a draft, which requires human-computer interaction and iteration to continually adjust and optimize the film content based on user feedback.
To address these issues, we introduce AutoDirector, an interactive multi-sensory composition framework that supports long shots, special effects, music scoring, dubbing, and lip-syncing. This framework improves the efficiency of multi-sensory film production through automatic scheduling and supports the modification and improvement of interactive tasks to meet user needs. AutoDirector not only expands the application scope of human-machine collaboration but also demonstrates the potential of AI in collaborating with humans in the role of a film director to complete multi-sensory films.
\end{abstract}

\section{Introduction}

With the advent of generative models, single-modality synthesis, such as music, visuals, and speech, has reached remarkable levels of realism. However, as illustrated in Figure \ref{fig:case}, multi-sensory composition, which aims to create cohesive and immersive videos by integrating various sensory elements into a unified whole, encompasses a more extended and intricate process than traditional video synthesis, involving scriptwriting, shooting, scoring, dubbing, and special effects~\cite{aldausari2022video,le2021ccvs}. The scarcity of readily available, clearly copyrighted resources for multi-sensory videos further complicates the direct training of models for this purpose.

While AI agents~\cite{reed2022generalist,andreas2022language,deng2024mind2web}, which leverage the collaboration of multiple models, have shown promising results in various fields, their application in multi-sensory video composition remains limited. Two significant challenges persist in this domain. First, existing AI agents~\cite{gronauer2022multi,wen2022multi,amirkhani2022consensus,hong2023metagpt} often struggle with time management in complex, time-consuming tasks like movie production. Without efficient scheduling and automatic planning, system speed can be severely hindered. Second, the iterative communication process between users and directors is crucial in movie production to ensure the final product meets user expectations. The lack of a robust interactive approach in current systems limits the expressive capabilities of multi-sensory films.

\begin{figure}[h!]
	\centering
	\includegraphics[width=8cm]{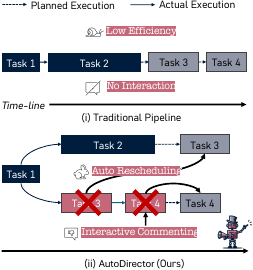}
	\caption{\textbf{Overview of cognitive scheduling.} Unlike traditional sequential execution, which is inefficient and unable to communicate user needs, our method can continuously organize and arrange tasks based on user comments and efficiently carry out movie creation through parallel execution.}
	\label{fig:intro}
\end{figure}

To address these challenges, we propose AutoDirector, an interactive agent system designed to handle tasks in parallel and engage actively with users. Unlike previous systems, AutoDirector dynamically plans tasks based on feedback, efficiently generating high-quality films by automatically scheduling activities such as scriptwriting, shooting, scoring, dubbing, and special effects. As shown in Figure \ref{fig:intro}, it also facilitates real-time user interaction to modify or enhance tasks according to user needs.

Our experimental results demonstrate that the AutoDirector model effectively overcomes existing challenges, advancing the field of multi-sensory video composition. This framework enhances production efficiency through automatic scheduling and supports iterative task modification based on user feedback. AutoDirector not only broadens the scope of human-machine collaboration but also showcases the potential of AI in assisting film directors to create multi-sensory films collaboratively.

Our contributions are three-fold:
\begin{itemize}
\item We present AutoDirector, an agent system capable of performing tasks in parallel and actively engaging with users. Unlike previous agent systems, AutoDirector dynamically schedules tasks based on user feedback.
\item We introduce the novel scenario named multi-sensory movie generation, successfully implemented with AutoDirector. AutoDirector employs scheduling strategies to automate the execution of modules and incorporate user feedback, thus optimizing task execution.
\item Experiments validate the effectiveness of AutoDirector, showing that it outperforms the baseline model in both efficiency and quality of multi-sensory composition.
\end{itemize}
\section{Related Work}

\paragraph{Video Generation}

Recently, the emergence of Sora~\cite{sora} has elevated video synthesis to new heights, capable of generating a minute of high-fidelity video. In the field of video synthesis, several related works have been conducted. PiKa~\cite{pika} provided an effective tool for editing videos through textual prompts. NUWA-XL~\cite{yin2023nuwa} presented a novel coarse-to-fine diffusion over Diffusion architecture for generating extremely long videos. Recently, thanks to the strong in-context learning capability of Large Language Models (LLMs), movie generation according to textual scripts has become a new direction. DirecT2V~\cite{hong2023direct2v} leveraged instruction-tuned LLMs as directors for text-to-video conversion, facilitating the inclusion of time-varying content and ensuring consistent video production. Vlogger~\cite{zhuang2024vlogger} used an LLM as Director, breaking down vlog creation into four stages, mimicking human roles, and employing a novel video diffusion model for scene generation. MovieLLM~\cite{song2024moviellm} employed GPT-4~\cite{achiam2023gpt} to create entire storylines by first generating descriptions for each frame, which were then used to produce high-quality, frame-by-frame videos based on provided instructions. However, generating videos frame by frame is very expensive, and it lacks smooth transitions, dubbing, and background music, making it unacceptable for human audiences. 

In this work, by facilitating detailed multi-turn interaction between users and the director, our model can generate more satisfying videos that closely resemble natural scenes.

\paragraph{Agent System}
LLMs~\cite{touvron2023llama,vicuna,gpt2,flant5} have been used to streamline and unify processes in various fields, such as the software development process~\cite{qian2023communicative}, recommendation systems~\cite{huang2023recommender,vullam2023multi}, medical treatment~\cite{calisto2023assertiveness}, etc. By incorporating agent systems, existing works integrated domain-specific expert models with LLMs to accomplish multiple tasks, mimicking human intelligence. The LLM-powered agent has diverse applications, ranging from gaming environments like Minecraft~\cite{wang2023voyager} to real-world scenarios~\cite{qin2023toolllm,wu2023autogen}. ToolLLM~\cite{qin2023toolllm} facilitated the integration of LLMs with various real-world APIs. 
Autogen~\cite{wu2023autogen} enabled conversations between multiple agents, enhancing the ability of LLM-based agents to tackle complex tasks. 

However, existing pipeline workflows cannot reasonably schedule tasks and interact with users in real time. Our work highlights the potential for significant speedups and improved efficiency by integrating these into pipelines.

\section{Methodology}

\begin{figure*}[h]
	\centering
	\includegraphics[width=17.5cm]{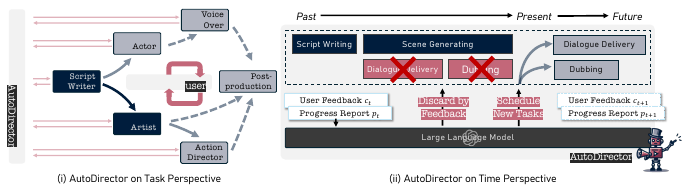}
	\caption{\textbf{The film production process of AutoDirector.} Our process can be interpreted from two different perspectives: the task perspective and the time perspective. From the task perspective, film production consists of a series of tasks, and there is a sequence between the tasks. The AutoDirector manages all the tasks, and the user continuously puts forward their requirements during this process. From the time perspective, the AutoDirector will get the current progress report and user feedback at the beginning of each time segment and arrange new tasks, revoke completed tasks, or wait based on this until completion.}
	\label{fig:method}
\end{figure*}

\subsection{Overall of AutoDirector}

\paragraph{Task Perspective}

In the left part of Figure \ref{fig:method}, different from video generation, for film production, we need the cooperation of multiple crew members, such as scriptwriters, actors, and voice actors, and each crew member will be responsible for a task, such as script creation, acting, and dubbing. In our work, we will use AutoDirector to arrange tasks. Not only that, there is also a user who constantly pays attention to the intermediate results and gives comments. We believe that for complex tasks like film production, users' opinions are crucial because this will ensure that the film is satisfactory to the director and the users.

\paragraph{Time Perspective}

In the right part of Figure \ref{fig:method}, unlike traditional linear pipelines, for movie production, there are networked dependencies between tasks. Some tasks can be done in parallel, while others must wait for other tasks to be completed before proceeding. In our work, the most important task of the AutoDirector, played by the Large Language Model (LLM), is to arrange the schedule of tasks continuously. Additionally, the AutoDirector will revoke certain tasks based on user feedback and task progress reports and reschedule certain tasks after improvement. After each round of progress reports and user feedback, the AutoDirector always tracks and plans tasks in a continuous loop. When there is nothing to do, the AutoDirector waits for the following progress report and user feedback until the movie production is completed.

\subsection{AutoDirector with Cognitive Scheduling}

In order to complete the film production, we have a series of tasks to complete. We call each task an event. We view film production as a sequence of events $\mathcal{E}$, containing $n$ events.

Some events, such as performances, must wait for previous events to be completed before they can begin, such as script creation. We define its dependent event set $\mathcal{D}$ for this.

Film production is a long process, and we divide time into continuous time slices $\{t_i\}^{T}_{i=0}$. The AutoDirector will continue to plan the time slices until all events are completed.

At the beginning of each time segment $t$, we will get the progress report $p_t$ of all tasks, which is composed of the list of events currently being executed and the events that have been completed. The AutoDirector can plan what to continue waiting for or arrange a crew member to execute an event in the next period through $p_t$.

At the same time, at the beginning of each time segment $t_t$, the user may provide feedback $c_t$. User feedback $c_t$ is the user's comments on the intermediate results, such as scripts, images, or dialogues.

If the user does not provide feedback, no event needs to be revoked, so $R_t = \emptyset$.

The AutoDirector will arrange the upcoming tasks:

\begin{equation}
    Q_t = f_{\mathrm{q}}(p_t; \mathcal{E}, \mathcal{D}).
\end{equation}

If the dependencies of multiple events are satisfied, $Q_t=\{e_k\}$ may contain multiple events.

In the absence of any dependencies being met, no events will be scheduled:

\begin{equation}
Q_i = \emptyset.
\end{equation}

However, if the user has feedback on some content, the AutoDirector will try to retry some of the steps. According to the progress report $p_t$ and user feedback $c_t$, we can get the list of events to be revoked:

\begin{equation}
R_t = f_{\mathrm{r}}(c_t, p_t; \mathcal{E}, \mathcal{D})
\end{equation}

At the same time, the AutoDirector will rearrange the started tasks $Q_t$ by:

\begin{equation}
    Q_t = f_{\mathrm{q}}(c_t; \mathcal{E}, \mathcal{D}).
\end{equation}

After confirming the revoked event $R_t$ and the planned event $Q_t$, the event will be executed as planned and attached to the user's feedback. From this, we can get the progress report for the next time period:

\begin{equation}
p_{t+1} = \phi(p_t; R_t, Q_t),
\end{equation}
where $\phi$ is the progress report update function.

Initially, $p_0$ does not contain any events. That is, there are events that have yet to be completed or events in progress. The division of time periods is not uniform. When any event is completed, regardless of whether there are parallel events in progress, we will regard this time as the end of the previous period and the beginning of the next period. We can get the final result when all events in the list $\mathcal{E}$ are completed.

\subsection{Film Crews}

The entire movie production process is participated in by the following roles, each responsible for a certain part of movie production. Through the scheduling and collaboration of AutoDirector, the film is eventually produced.

\textbf{Scriptwriter.} The scriptwriter generates several scenes based on a requirement, outlining the characters involved and the plot for each scene.

\textbf{Artist.} The art director designs the scenes, actors' states, costumes, and makeup for each scene, relying on the screenwriter's concept for each scene and character. The art director uses PixelArt \cite{chen2023pixartalpha} as a tool to generate characters from text.

\textbf{Actors.} The actors improvise the specific dialogues for each scene, relying on the screenwriter's plot design.

\textbf{Action Director.} The action director shoots the specific actions for each scene and must wait until the art direction is completed before proceeding. The action director uses Stable Video Diffusion \cite{blattmann2023stable} as a tool to generate long shots from keyframes.

\textbf{Voice Over.} The voice over dubs the dialogue for each scene and needs to wait until the actors finish their performances before they can start dubbing. We use Azure Text2speech as a dubbing tool.

\textbf{Post-production.} The post-production process the results of each scene's shooting with special effects and add dubbing and music. Before proceeding, they must wait until the dubbing, action, and art direction are completed. For dubbing and music synthesis, we call Pika's Lip Sync feature, and MusicGen \cite{copet2023simple} to generate music.

\textbf{User.} Unlike the crew members played by LLM mentioned above, we also include a human user who will interact with the director in a timely manner to meet their needs. In any round, the user can also choose not to comment.

\subsection{Implementation Details}

AutoDirector is a GPT-4-based agent system. For Script Writing, it delegates the task to another agent. AutoDirector works with other tool modules through API calls to complete the remaining steps, such as video synthesis. AutoDirector only needs to run on a CPU. However, the tools we call in this paper will consume additional resources. To ensure the visual synthesis model works properly, we used Nvidia A100 80GB GPU in our experiments. All random seed is fixed into $42$.

\paragraph{Emotion-aware Dubbing}
Using LLM, each line of dialogue will be assigned a unique emotion, such as anger, surprise, or whisper. Emotion will serve as a parameter when invoking Text2speech to synthesize voiceovers that match the characters' emotions.

\paragraph{Video Synthesis}
A movie comprises multiple long shots, each representing a scene. However, the length of each shot is different due to the changes in dialogue and plot. For each scene, the art task will first generate a scene frame as the scene. Next, we will use the scene frame as a condition to call Stable Video Diffusion to generate extended frames. Since the length generated by Stable Video Diffusion is fixed, we will use the end of the generated frame as the Key Frame for extension and further iterative generation. To solve the error accumulation problem, we will add a set of reverse frames after the first round of extension so that the first keyframe is consistent with the scene frame. Our long shots will match the length of the voiceover.

\paragraph{Theme-aware Background Music}
AutoDirector will also generate corresponding background music based on the whole play. It is worth noting that AI also synthesizes the background music.

\paragraph{Editing and Production}
The final film will splice multiple scenes together, and there will be cross dissolves between shots as transition effects based on the template. Background music and voiceovers will be merged into a unified sound channel to achieve a cinematic audiovisual effect.

Please refer to the \textbf{Supplementary Materials} for more details on implementation.
\section{Experiment}

\subsection{Experiment Setup}

This paper proposes a novel scenario, multi-sensory composition, which aims to create cohesive and immersive videos by integrating various sensory elements into a unified whole. To evaluate the effectiveness of AutoDirector in multi-sensory composition, we selected several movie themes as inputs and manually assessed the generated results across three dimensions: visual aesthetics, narrativity, and controllability.
For visual aesthetics, we evaluated the aesthetic quality of individual video frames and the coherence across multiple frames. In terms of narrativity, we assessed the coherence, logic, and emotional expression of the story. Controllability was measured by the degree of user control over the generated video.

To ensure a fair comparison, we established a baseline model that utilizes the same tools as AutoDirector, such as video synthesis and music synthesis models, but lacks the planning and interaction capabilities of AutoDirector.

We also included NUWA-XL, Vlogger, OpenSora, and Pika as references. However, it is important to note that these models are traditional video synthesis tools that only support video generation and do not align with the goal of multi-sensory movie generation.

\subsection{Overall Results}

\begin{table*}[tb]
  \caption{\textbf{Overall results of quality.} AutoDirector stands out among all models for its unique ability to finely control movie generation, producing aesthetically pleasing images and high-quality scripts, and improving narrative logic, thus significantly outperforming other models in the overall score. 
  }
  \label{tab:overall}
  \setlength{\tabcolsep}{10.5pt}
  \centering
  \begin{tabular}{lccccccc}
    \toprule
    {\textbf{Methods}}  & {Vision} & {Sense} & {Audio} & {{Aesthetics}} &{{Narrativity}} &{{Controllability}} &{{Overall}}\\
    \midrule
    NUWA-XL & $\checkmark$ &  & &24 & 10 & 15 & 16.4 \\
    Vlogger & $\checkmark$ &  & &38 & 58 & 29 & 42.3 \\
    OpenSora & $\checkmark$ &  & &61 & 27 & 30 & 39.8 \\
    Pika & $\checkmark$ & & &\textbf{86} & 36 & 54 & 58.9 \\
    \midrule
    Baseline (Ours) & $\checkmark$ & $\checkmark$& $\checkmark$& 57 & 48 & 22 & 43.4\\
    \textbf{AutoDirector (Ours)} & $\checkmark$ & $\checkmark$& $\checkmark$& \textbf{86} & \textbf{84} & \textbf{83} & \textbf{84.4}  \\
  \bottomrule
  \end{tabular}
\end{table*}

\begin{table}[tb]
  \caption{\textbf{Overall results of efficiency.} AutoDirector can significantly enhance the generation efficiency compared to the pipeline method.
  }
    \setlength{\tabcolsep}{1pt}
  \label{tab:time}
  \centering
  \begin{tabular}{lcccc}
    \toprule
    \textbf{Methods} & Parallel & Interactive & Time$^{\downarrow}$ & Multiple$^{\downarrow}$ \\
    \midrule
    Baseline (Ours) &&& 226.07 & 1.00$\times$\\
    \textbf{AutoDirector (Ours)} & $\checkmark$ & $\checkmark$ & \textbf{138.71} & \textbf{0.61}$\times$ \\
  \bottomrule
  \end{tabular}
\end{table}

\paragraph{Effectiveness}

As shown in Table \ref{tab:overall}, AutoDirector is capable of controlling movie generation at a fine-grained level, significantly improving controllability. By interacting with users to obtain human-perspective information, AutoDirector can generate scenes that better align with human aesthetics (kindly refer to Section \ref{sec:int}), thereby achieving much higher aesthetic scores compared to the baseline.

In terms of narrativity, AutoDirector can generate scripts of extremely high quality. This is because users can interactively evaluate and improve scripts with AutoDirector to address any narrative logic defects. This is also a key reason why AutoDirector significantly outperforms the baseline model in the overall score.

Since existing models do not support multi-sensory video generation, we have referenced some traditional video generation models for comparison. In theory, AutoDirector supports any video synthesis model as the backend. In this paper, as a demonstration, we used the open-source Stable Video Diffusion as the video synthesis model because it allows precise control over calls and timing, facilitating better quantitative experiments.

\paragraph{Efficiency}

AutoDirector exhibits excellent multi-sensory video generation effects and significantly reduces generation time. Since the backbone speed of existing models varies, for a fair comparison, we selected the version of AutoDirector without the scheduling mechanism as the baseline model, where all tasks are executed in series. Additionally, in this experiment, we assumed that there was no feedback from the user. As shown in Table \ref{tab:time}, AutoDirector significantly improves generation speed, with an acceleration of about 40\%. This is because many steps in movie production can be executed in parallel. Through reasonable module scheduling, AutoDirector can accelerate the generation of movie videos without any performance degradation. We can find that our method has the ability to reasonably plan tasks, execute tasks in parallel, or wait for tasks to be completed at appropriate times. Furthermore, our method can also revoke completed tasks based on user feedback and effectively transform them into new task goals to correct the tasks. By explicitly participating in the production process, users have relatively higher satisfaction with the results.

\subsection{Ablation Studies}

In order to verify whether each part of AutoDirector achieves the expected effect, we conduct ablation experiments on two different sub-modules: time scheduling and user interaction in Table \ref{tab:abl}. After removing the time scheduling, we found that although the effect did not change significantly, the user's satisfaction with time efficiency significantly decreased, reflecting the important role of time scheduling. At the same time, without user interaction, both time and effect satisfaction show a significant decrease. This is because, without intermediate communication, users will amplify their perception of time and tend to think that the model's execution time is longer. Moreover, without effective communication on intermediate results, satisfaction with the final result drops significantly, which proves the importance of interaction in complex tasks. By combining the two, AutoDirector achieves the best state regarding both time and effect satisfaction.

\begin{table}[tb]
  \caption{\textbf{Ablation results.} Both time scheduling and user interaction play indispensable roles, and their combination results in the highest user satisfaction.
  }
  \label{tab:abl}
  \centering
\setlength{\tabcolsep}{6.7pt}
  \begin{tabular}{lccc}
    \toprule
    \textbf{Methods} & Time$^{\uparrow}$ & Effect$^{\uparrow}$ & Overall$^{\uparrow}$ \\
    \midrule
    \textbf{AutoDirector (Ours)} & \textbf{72} & \textbf{74} & \textbf{73.0}\\
    \quad w/o Time Scheduling & 52 & 69 & 60.5 \\
    \quad w/o User Interaction & 37 & 7 & 22.0 \\
  \bottomrule
  \end{tabular}
\end{table}

\subsection{Discussion of Interaction}
\label{sec:int}
\begin{figure*}[h]
	\centering
	\includegraphics[width=17.5cm]{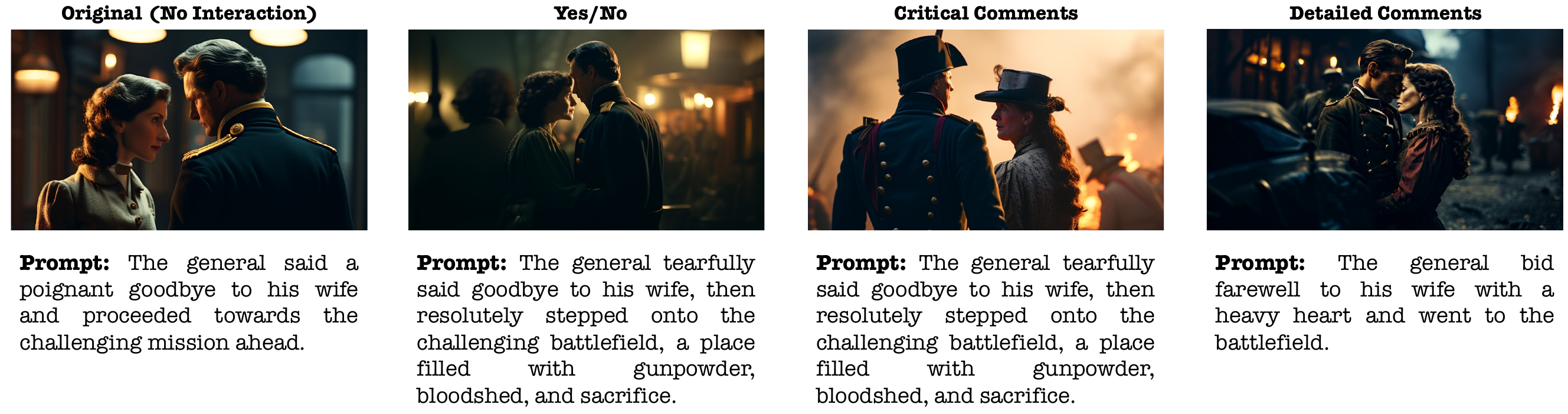}
	\caption{\textbf{Comparison of different interactive types.} The level of user feedback, ranging from \texttt{Yes/No} answers to \texttt{Detailed Comments}, progressively enhances the expressiveness of the picture, with a higher degree of participation leading to a more nuanced and emotionally impactful final product.}
	\label{fig:comp}
\end{figure*}

\subsubsection{Comparison of Interaction Frequency}

\begin{table}[tb]
  \caption{\textbf{Comparison of interaction frequency.} Increased interaction leads to improved satisfaction in both the intermediate and final results. Unlimited interaction greatly enhances the final result by interrupting error accumulation.
  }
  \label{tab:inter}
  \centering
\setlength{\tabcolsep}{9.7pt}
  \begin{tabular}{lccc}
    \toprule
    \textbf{Frequency} & Global$^{\uparrow}$ & Result$^{\uparrow}$ & Overall$^{\uparrow}$ \\
    \midrule
    None (Baseline) & 37 & 7 & 27.0 \\
    Low & 60 & 29 & 45.5 \\
    Intermediate & 75 & 59 & 67.0 \\
    No Limits & \textbf{76} & \textbf{74} & \textbf{75.0} \\
  \bottomrule
  \end{tabular}
\end{table}

To verify the importance of interaction in movie production, we designed a set of comparisons to observe the satisfaction of users with the intermediate and final results under different interaction frequencies. As shown in Table \ref{tab:inter}, we find that when no interaction occurs, the satisfaction of users with the intermediate process and the final result is lower. When a low frequency of interaction is added, the satisfaction with the intermediate process is significantly improved, but the improvement in satisfaction with the final results is limited because the low frequency of interaction is not enough to affect the final result. When the interaction frequency is further increased, the satisfaction with both the intermediate process and the final results is significantly improved. When unlimited interaction is allowed, although the satisfaction of the intermediate process is not significantly improved, the final result is more likely to satisfy the user because the accumulation of errors is interrupted.

\subsubsection{Comparison of Interaction Types}

How does the interaction method affect movie production? To explore this question, we designed a few different interaction methods. \texttt{Yes/No} represents that users only give feedback on whether they agree with the current effect and do not make suggestions. \texttt{Critical Comments} mean that users will point out dissatisfying content but do not provide their own thoughts. \texttt{Detailed Comments} represent that users will not only point out dissatisfying content but also provide their own ideas. In Figure \ref{fig:comp}, we find that even if only \texttt{Yes/No} evaluations are carried out, the expressiveness of the picture is significantly improved compared to the original picture because the details of the picture background are significantly enhanced. Using \texttt{Critical Comments}, the model further highlights the cruelty of the war. For pictures using \texttt{Detailed Comments}, the picture not only highlights the tension before the war but also emphasizes the reluctance of the lovers, significantly enhancing the expressiveness of the picture. This also proves that the level of user participation affects the final effect.

\subsection{Case Studies}

\subsubsection{Agents Workflow}

\begin{figure*}[h]
	\centering
	\includegraphics[width=17.5cm]{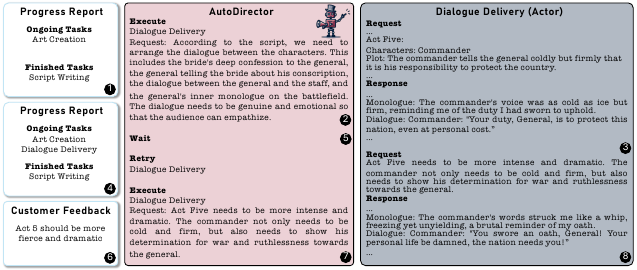}
	\caption{\textbf{Movie production process involving a user comment. } The process is dynamic, with decisions being made based on task process and user feedback, leading to adjustments and improvements in tasks such as dialogue generation.}
	\label{fig:process}
\end{figure*}

Figure \ref{fig:process} illustrates a segment of the complete movie production process. The numbers in the figure denote the sequence of events: the left side represents the progress report or user feedback, the middle shows the decisions made by the AutoDirector, and the right side depicts the actor agent executing the dialogue delivery. Some prompts have been omitted to convey the process clearly.

In Event \ding{182}, the progress report indicates that script writing has been completed and that art creation is underway. In the subsequent Event \ding{183}, the AutoDirector determines that dialogue delivery can proceed simultaneously, as its prerequisite tasks are complete. Therefore, it issues instructions to the actor and outlines the main objectives of the dialogue. In Event \ding{184}, the actor begins executing this command. Then, in Event \ding{185}, a new progress report is generated. Notably, because long-text generation takes time, Event \ding{184} is still ongoing when Event \ding{185} occurs. Consequently, in Event \ding{186}, the AutoDirector decides to wait, as no new commands can be executed at that moment. Once Event \ding{184} is completed, it is presented to the user.

At the time of Event \ding{187}, the user reviews the generated dialogue but is dissatisfied and suggests improvements for the fifth act. In Event \ding{188}, the AutoDirector considers the user's feedback, revokes the dialogue task, and re-executes it with the new requirements. In Event \ding{189}, the actor successfully adjusts the dialogue according to the user's specifications.

\subsubsection{Multi-sensory Movie Generation Result}

As shown in Figure \ref{fig:case}, we present an example of the final movie production effect of AutoDirector: a 1 minute and 18 seconds movie titled ``\texttt{The General's Wedding}", which tells the story of a general being drafted on his wedding day. The movie includes images, dialogues, and background music, presented here through frame images and text. AutoDirector generated a coherent series of plots and shots, crafting a vivid story with scenes containing multiple characters.

We observe that AutoDirector adds tone descriptions to the dialogue, enabling AI to generate voiceovers in different scenes and styles. When combined with lip sync, the final effect of the movie closely resembles that of a human-made film. Furthermore, the AI-generated background music, with its melancholy and heroic tune, enhances the romantic and tragic atmosphere of the movie. Through these examples, we demonstrate the effectiveness of AutoDirector and explore the potential of AI in movie creation.

\section{Broader Impact and Limitations}

AutoDirector's capability of fine-grained control over movie generation opens up new possibilities for content creation. Integrating human interaction and time scheduling improves the quality of the generated content and significantly increases efficiency. This can greatly enhance the film production process, reduce costs, and enable more creative ideas to be realized.
While AutoDirector improves the efficiency of movie generation, it still demands a certain amount of computational resources, which may limit its applicability for low-resource settings.

\section{Conclusion}

In this paper, we have introduced AutoDirector, an innovative interactive multi-sensory composition framework designed to address the challenges of generating multi-sensory outputs for high-value scenarios such as film directing. AutoDirector effectively tackles two major challenges: the lack of parallelism and online scheduling in production steps, as well as the diverse and evolving needs of users during the filmmaking process. By incorporating automatic scheduling and interactive modification capabilities, AutoDirector enhances the efficiency of multi-sensory film production and enables continuous adjustment and optimization based on user feedback. This framework not only broadens the scope of human-machine collaboration but also showcases the potential of AI in assisting humans in the role of film directors to create complex multi-sensory films.

\bibliography{aaai25}

\end{document}